# Embodied Self-Supervised Learning (EMSSL) with Sampling and Training Coordination for Robot Arm Inverse Kinematic Model Learning

Qu Weiming[&], Liu Tianlin[&], Wu Xihong, Luo Dingsheng[*]
National Key Laboratory of General Artificial Intelligence, Key Laboratory of Machine Perception (MoE),
School of Intelligence Science and Technology, Peking University, Beijing 100871, China.
Email: {quweiming, liutl, xhwu, dsluo}@pku.edu.cn
[&]These authors contributed equally to this work and should be considered co-first authors.
* Corresponding author

***Abstract*— Forward and inverse kinematic models are fundamental to robot arms, serving as the basis for the robot arm's operational tasks. However, in model learning of robot arms, especially in the presence of redundant degrees of freedom, inverse kinematic model learning is more challenging than forward kinematic model learning due to the non-convex problem caused by multiple solutions. Besides, Current learning-based methods often segregate data sampling from model training, potentially leading to suboptimal data utilization and restricted model adaptability. In this paper, we introduce the concept of "Embodiment" and propose a framework for autonomous learning of the robot arm inverse kinematic model based on embodied self-supervised learning (EMSSL) with sampling and training coordination, effectively solving the non-convex problem of the inverse kinematic model and significantly enhancing data sampling efficiency. Concurrently, we investigate batch inference and parallel computation strategies for data sampling to expedite model learning. Additionally, we develop two approaches for the fast adaptation of the robot arm models. A series of experimental evaluations attest to the efficacy of our proposed method.**

## I. INTRODUCTION

Robot arms have gained significant importance in recent years in various fields, such as industrial manufacturing, medical surgery, home services, and space exploration. For robot arm manipulation, the two fundamental robot arm kinematic models are the forward kinematic model and the inverse kinematic model. The forward kinematic model is to compute the position and orientation of the tool frame relative to the base frame, given a set of joint angles, while the inverse kinematic model is to calculate the joint angles that could be used to attain a given position and orientation of the end-effector [1]. However, the inverse kinematic model learning is often more complex, especially in the presence of redundant degrees of freedom, as redundant degrees of freedom may cause the inverse kinematic model to have multiple solutions, i.e., the same end position corresponds to multiple solutions in the joint space. These solutions may form a non-convex solution space and lead to a non-convex problem [2].

In addition, for learning based methods of modeling the inverse kinematics, how to sample the training data is a key issue. Currently, common methods include Motor Babbling [3][4], Goal Babbling [5][6], and Constrained DOF Exploration [7][8]. However, for inverse kinematic model learning, one of the main problems with the current methods is the separation between data sampling and model training, i.e., the data is first collected and then used to train the model. This separation suffers from poor data utilization and limited model adaptability, since if the robot arm geometrical or dynamic parameters change during use due to mechanical wear, all the data needs to be sampled again. Unfortunately, data acquisition is often difficult and costly for robots, especially for a robot arm with high degrees of freedom. To address this separation, Sun et al. [9] proposed an embodied self-supervised learning method and illustrated its effectiveness by solving the problem of acoustic-to-articulatory inversion.

In this paper, drawing on the embodied self-supervised learning method previously used in speech inversion [9], we propose a framework for robot arm inverse kinematic model based on embodied self-supervised learning (EMSSL) with sampling and training coordination (Section III-A), where data sampling and model training iteratively proceed and promote each other, similar to the process of Boosting in machine learning [10] and the "guess-try-feedback" process in human learning [11]. (Since EMSSL is a learning-based method, for the sake of simplicity in expression, we will use the term "model" as a substitute for the "kinematic model".) Unlike conventional self-supervised learning, EMSSL is under the constraint of a physical model, which utilizes this physical model as a guide for updating the inverse model, effectively addressing the non-convex problem inherent in the inverse model. In this paper, the forward model of the robot arm is the physical model in EMSSL. Furthermore, we acknowledge that the original EMSSL method exhibits a relatively slow pace during the data sampling phase. To address this, we propose an optimized acceleration strategy based on inverse model batch inference and forward model parallel computation (Section III-B). Lastly, we recognize that long-term usage of the robot can lead to mechanical wear and tear, which in turn results in a certain degree of deviation in the previously trained model. At the same time, due to the gap between the simulation environment and the real environment, models applicable in the simulation environment may not be directly usable in the real environment. Therefore, the fast adaptation for robot arm models is essential. Drawing inspiration from the characteristics of human learning, particularly the ability to reuse and adapt previously acquired knowledge to new situations, we develop two fast adaptation approaches for the robot arm model under the EMSSL framework (Section III-C).

In summary, the main contributions of this paper are summarized as follows:

- We propose an EMSSL framework with sampling and training coordination for robot arm inverse model learning and investigate an optimized acceleration strategy based on inverse model batch inference and forward model parallel computation, greatly improving data sampling efficiency and model convergence rate.
- We introduce the concept of "Embodiment", effectively solving the non-convex problem of the inverse model through the constraints of the robot arm's forward model in the EMSSL framework.
- Drawing from the characteristics of human learning, namely the reuse and adaptation of previously learned knowledge to generalize to new situations, we develop two fast adaptation approaches for the robot arm model under the EMSSL framework.

## II. RELATE WORKS

### A. Robot Arm Inverse Model Learning

Traditional methods used for robot arm inverse model learning include Analytical Solution Methods [12] and Numerical Solution Methods [13][14]. Analytical solutions are derived in closed form, but they depend on the structure of the robot and need to be derived independently for each robot. Numerical Solution Methods adopt an iterative approach, which is usually slower but more general than Analytical Solution Methods, and do not depend on a specific structure of robots. However, Numerical Solution Methods are sensitive to the initial conditions of the iterations and do not always guarantee convergence to a correct solution.

Aiming at the shortcomings of traditional methods, learning-based methods have been widely used. Almusawi et al. [15] proposed a new inverse model learning architecture based on Feedforward Neural Network (FNN), which takes both the desired end position and the current joint angle information of the robot arm as input. Elkholy et al. [16] proposed the use of Convolutional Neural Networks (CNN) to learn the inverse model. In recent years, reinforcement learning methods have also been used for inverse model learning. Phaniteja et al. [17] used Deep Deterministic Policy Gradient (DDPG) to dynamically generate joint angle solutions for a humanoid robot. Blinov et al. [18] proposed a deep Q-Learning algorithm for solving the inverse model of a four-linked robot arm. Malik et al. [19] adopted Deep Q-Network (DQN) to solve the inverse kinematics problem of a 7-degree-of-freedom robot arm. In addition, some researchers have also used genetic algorithms to learn inverse models [20-22]. However, the accuracy of inverse models obtained by reinforcement learning methods is usually poor, while genetic algorithms are computationally intensive and slow to converge.

### B. Approaches to the non-convex problem

Due to the existence of the non-convex problem, direct regression is not suitable for inverse model learning, because direct regression tends to average the solutions in the non-convex solution space. Unfortunately, the average result can be invalid, which leads to poor performance [23]. To address this problem, various methods have been proposed, including Configuration Space Splitting [24][25], goal babbling [5][26], distal supervised learning [27][28], and joint distribution modelling [29-34].

**Configuration Space Splitting** [24][25] is similar to the partitioning strategy, which is to partition the entire configuration space (joint space) into multiple subregions so that the global non-convex problem can be transformed into multiple local convex problems, and then the global solution can be constructed from these local solutions. However, a database is usually required to decide which local model should be selected, and it can be difficult to obtain such a database in a high-dimensional space.

**Goal Babbling** [5][26] involves learning the inverse model by goal-directed sampling along a path at a given position, and discarding redundant solutions. Therefore, two issues need to be considered: the specific sampling scheme and the strategy for discarding redundant solutions. If the redundant solutions are discarded improperly, the solutions at neighboring positions of an expectation position may also cause a non-convex problem.

**Distal Supervised Learning** [27][28] learns the forward model and the inverse model as a whole, and updates the inverse model by gradient back propagation. Essentially, Distal Supervised Learning is also goal-oriented, which learns a particular solution among multiple valid solutions. The main shortcoming of this method is that the forward model needs to be derivable because the inverse model learning needs to back propagate the gradient with the help of the forward model. Moreover, the performance of the inverse model will be affected by the accuracy of the forward model.

**Joint Distribution Modelling** [29-34] directly learns the joint distribution of multiple solutions, such as Mixture Density Network (MDN) [29], Invertible Neural Network (INN) [30][31], and Conditional Generative Adversarial Network (CGAN) [32-34]. The problem with these approaches is that although the overall distribution of multiple solutions can be learned, the accuracy of the obtained models is generally low.

### C. Approaches to data sampling

For an autonomous robot, the ability to actively explore the environment and obtain training data is important [35]. Different approaches are used for this purpose, including Motor Babbling [3][4], Goal Babbling [5][6], and Constrained DOF Exploration [7][8].

**Motor Babbling** [3][4] obtains training data by random joint movements, requiring a search of the entire joint space and therefore has high spatial complexity. **Goal Babbling** [5][6], in contrast, searches in the task space by selecting a position as a target and obtaining training data by continuously trying to reach that target. Baranes et al [36] demonstrated that, when the target point is properly selected, Goal Babbling generates enough training data for inverse model learning faster than Motor Babbling. **Constrained DOF Exploration** [7][8] explores one joint at a time, followed by exploration of the next joint, while the other joint states remain unchanged. Compared with Motor Babbling and Goal Babbling, Constrained DOF Exploration has lower spatial complexity, as only one joint is explored at a time. However, it may ignore the training data that requires multi-joint joint search to obtain.

## III. METHODOLOGY

### *A. Embodied Self-supervised Learning*

The framework for inverse model embodied self-supervised learning is illustrated in Fig. 1. Given that we are focused on learning the inverse model, we assume that the forward model is already known. The learning process for the inverse model is carried out through embodied self-supervision: data is sampled in each iteration and then used for training. This process is essentially goal-oriented learning, i.e., learning with the desired position of the robot arm as the goal, and solving the non-convex problem of the inverse model through the constraints of the forward model.

During the data sampling phase, the inverse model deduces the joint angle required to reach the desired end position, and the forward model predicts the end position corresponding to that joint angle. The resulting data pair (joint angle, end position) is used for training. Unlike Distal Supervised Learning, our method does not require the forward model to be derivable, as it is only used to generate the data and does not need to help the inverse model to back propagate the gradient. In the model training phase, the predicted end position obtained from the data sampling phase is fed back to the inverse model to obtain a new joint angle. The joint angle from the data sampling phase is used as supervised information for the new joint angle, and the inverse model is updated using the gradient descent method. We use the minimum mean square error as the loss function to minimize the error of the reconstructed joint angle, and the loss function is as follows:

$$L(\boldsymbol{\theta}) = \frac{1}{N}\sum_{i=1}^{N}\left(\boldsymbol{q}^{(i)} - IM\left(\boldsymbol{p}^{(i)}\right)\right)^2 \quad (1)$$

where $IM$ denotes the inverse model, $\boldsymbol{\theta}$ is the parameter of the inverse model, and $\{(\boldsymbol{q}^{(i)}, \boldsymbol{p}^{(i)})\}_{i=1}^{N}$ is the dataset obtained in the data sampling phase.

The coordination between data sampling and model training involves an iterative relationship (as shown in Fig. 2), rather than a sequential one. The entire learning process will go through several iterations of data sampling and model training. The inverse model will be optimized in each iteration, and the optimized inverse model will be used for sampling in the next iteration. The new data obtained from sampling will be used for retraining the inverse model. After several iterations, the inverse model is considered to have converged when the outputs of the left and right inverse model are similar.

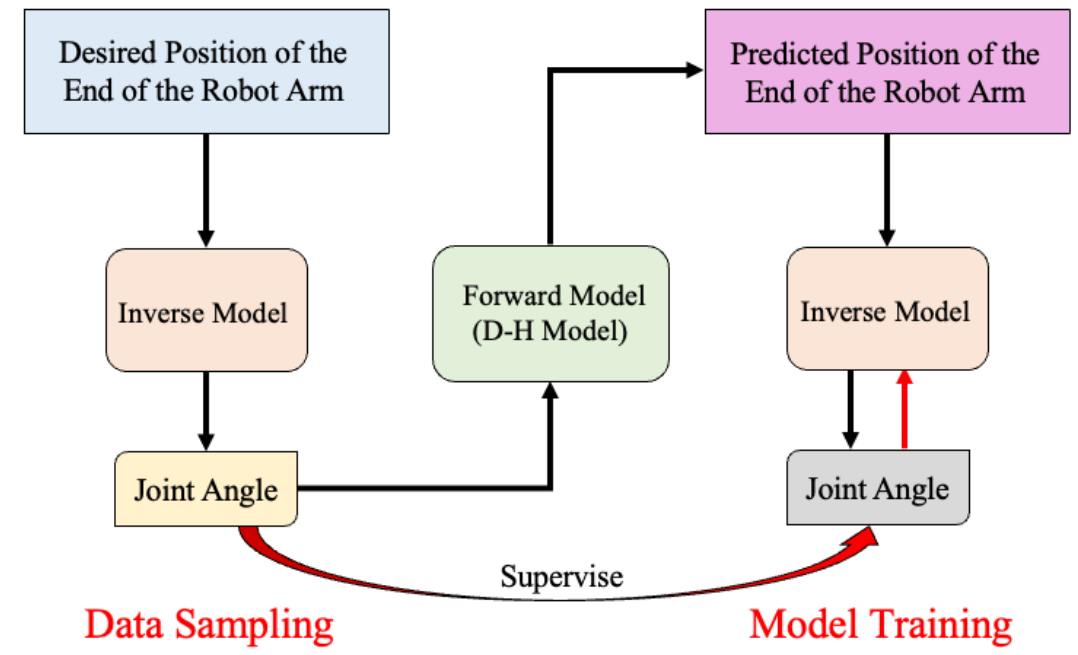

Fig.1. The framework of embodied self-supervised learning for robot arm inverse kinematic model learning. The left and right inverse models represent the same inverse model, while the left and right joint angles are different. The left joint angle is obtained based on the desired position of the robot arm end, while the right one is obtained based on the predicted position of the robot arm end. The left joint angle is the supervised information of the right joint angle.

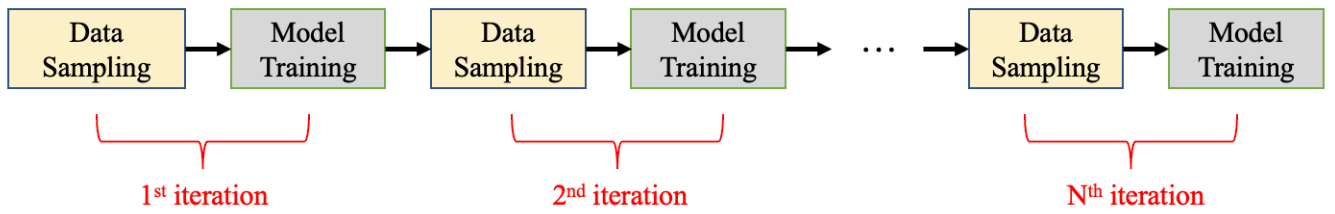

Fig.2. The schematic diagram of the coordination between data sampling and model training. Data sampling and model training iteratively proceed and promote each other, and as the iterations continue, the sampled data becomes better and better, and the inverse model becomes more and more accurate.

### *B. Data Sampling Accelerating*

Since for each data in the dataset, inverse model inference and forward model computation are performed sequentially, the data sampling phase can be time-consuming, resulting in slower learning of the inverse model. In this section, we propose an accelerated strategy based on inverse model batch inference and forward model parallel computation. This strategy involves inferring all the data in batch with the inverse model, followed by performing parallel computations uniformly with the forward model, as shown in Fig. 3.

The accelerated inverse model EMSSL algorithm is described in Algorithm 1. Specifically, multiple desired positions of the end of the robot arm are collectively inputted into the inverse model for batch inference, yielding a set of joint angles (as depicted in lines 4 ~ 9 of Algorithm 1). Subsequently, the forward model is enabled to execute parallel computations via multithreading, and by combining the joint angles derived from the batch inference of the inverse model with the predicted positions of the end of the robot arm obtained from the parallel computation of the forward model, the dataset is generated (as illustrated in lines 10 ~ 17 of Algorithm 1). The process of model training is as shown in lines 19 ~ 25 of Algorithm 1. The predicted positions of the end of the robot arm obtained during the data sampling phase are re-inputted into the inverse model. After re-inference by the inverse model, new joint angles are obtained. The joint angles inferred during the data sampling phase serve as the supervisory information for the new joint angles, and the parameters of the inverse model are updated using gradient descent.

**Algorithm 1** Accelerated Inverse Model EMSSL

**Input:** Forward Model FM, Inverse Model IM, Unlabeled dataset of robot arm end positions $\mathcal{U} = \{\boldsymbol{p}^{(i)}\}_{i=1}^{N}$, Maximum number of iterations $T$, Epochs $E$, Number of batches in a round of epoch for dataset (for example $\mathcal{X}$) $N_{B\mathcal{X}}$, Number of small batch samples for inference $M_R$, Number of small batch samples for training $M_T$, Number of parallel computation threads $K$, Learning rate $\eta$

1: Initialize the parameters $\boldsymbol{\theta}$ of the inverse model IM randomly
  Initialize the sample dataset: $\mathcal{D} \leftarrow \emptyset$
  Initialize the joint angle dataset of IM inference: $\mathcal{Q} \leftarrow \emptyset$
  Initialize the end position dataset of FM computation: $\mathcal{P} \leftarrow \emptyset$
2: **for** $t = 1 \ldots T$ **do**
3:   ** ***Data Sampling*** **
4:   Empty the dataset $\mathcal{D}, \mathcal{Q}, \mathcal{P}$
5:   **for** $n = 1,2 \ldots N_{B\mathcal{U}}$ **do**
6:     Sample small batches from $\mathcal{U}$: $\mathcal{B} \leftarrow \{\boldsymbol{p}^{(m)}\}_{m=1}^{M_R}$
7:     IM batch inference: $\mathcal{Q}_n \leftarrow IM(\mathcal{B})$
8:     Update the joint angle dataset: $\mathcal{Q} \leftarrow \mathcal{Q} \cup \mathcal{Q}_n$
9:   **end for**
10:   $j \leftarrow 0$
11:   **repeat**
12:     According to the number of threads $K$, pop the joint angles in order: $\mathcal{Q}_j = \{q_{1+j}, q_{2+j}, \ldots, q_{K+j}\}$
13:     FM parallel computation: $\mathcal{P}_j = FM(\mathcal{Q}_j)$
14:     Update the end position dataset: $\mathcal{P} \leftarrow \mathcal{P} \cup \mathcal{P}_j$
15:     $j = j + 1$
16:   **until** the data in $\mathcal{Q}$ have been computed
17:   Update the sample dataset: $\mathcal{D} \leftarrow \{\mathcal{Q}, \mathcal{P}\}$
18:   ** ***Model Training*** **
19:   **for** $\boldsymbol{e} = 1 \ldots E$ **do**
20:     **for** $n = 1,2 \ldots N_{B\mathcal{D}}$ **do**
21:       Sample small batches from $\mathcal{D}$:
        $\mathcal{B} \leftarrow \{(\boldsymbol{q}^{(m)}, \boldsymbol{p}^{(m)})\}_{m=1}^{M_T}$
22:       $L(\boldsymbol{\theta}) = \frac{1}{M_T} \sum_{i=1}^{M_T} \left(\boldsymbol{q}^{(m)} - IM(\boldsymbol{p}^{(m)})\right)^2$
23:       Update $\boldsymbol{\theta}$ with GD: $\boldsymbol{\theta} \leftarrow \boldsymbol{\theta} - \eta \nabla_{\boldsymbol{\theta}} L(\boldsymbol{\theta})$
24:     **end for**
25:   **end for**
26: **end for**

### C. *Fast Adaptation for Robot Arm Models*

In practical use scenarios, it is often necessary for robots to have the ability to quickly adapt their models for several reasons. First, due to the gap between the simulation environment and the real environment, models that are applicable in the simulation environment may not be directly usable in the real environment, necessitating model adaptation in the real environment. Second, the environment is dynamically changing, and there may be certain mechanical wear and tear in the robot arm during long-term use, which can cause certain deviations in the previously trained model, requiring model adaptation to correct these deviations. Third, compared with learning completely from scratch, model adaptation approaches can quickly obtain usable models, which is consistent with the characteristics of human learning, i.e., reusing and adapting previously learned knowledge and generalizing it to new situations.

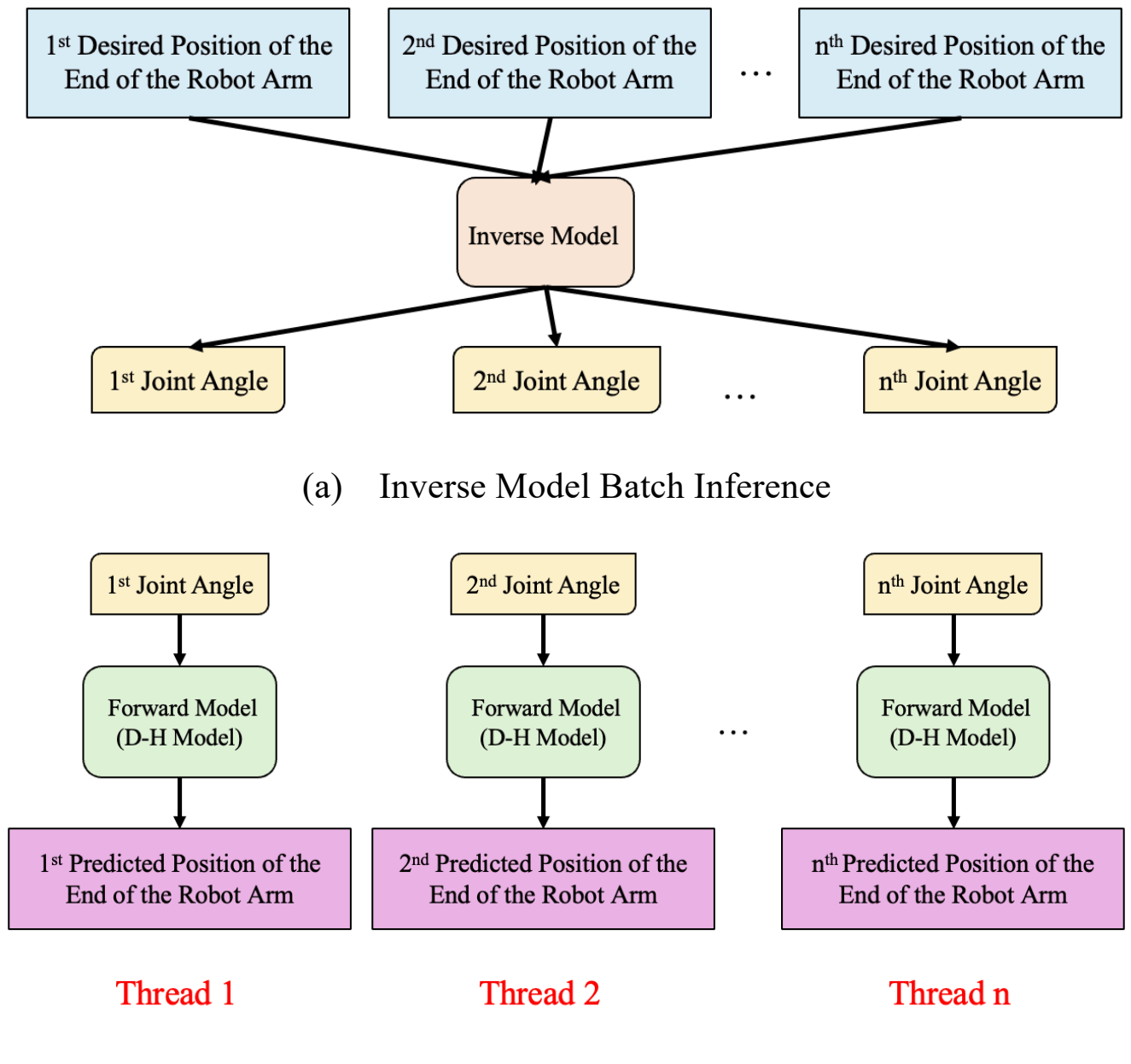

(b) Forward Model Parallel Computation

Fig.3. The schematic diagram of inverse model batch inference and forward model parallel computation. Unlike the previous algorithm, which requires sequential inference and computation for each data point in the dataset, the improved algorithm involves inferring all the data in batch with the inverse model and then uniformly performing parallel computation with the forward model.

In this section, we propose two approaches for fast adaptation for robot arm models. The assumed scenario here is that, according to the previously proposed framework (Fig.1), the inverse model of the robot arm has been trained, but now the robot arm has changed, and it is necessary to quickly adjust the inverse model. The first approach is shown in Fig. 4, where a small sample of data is sampled for the changed robot arm, and the forward model is updated by learning the link parameters by gradient descent. However, this causes the inverse model to become non-convergent. To achieve model adaptation, we propose several iterations of inverse model EMSSL with sampling and training coordination. The second approach, shown in Fig. 5, directly replaces the forward model with the real robot arm. Unlike the first approach, this approach does not require any link parameters, as the forward model is implicitly represented by the robot arm itself, which implicitly takes into account various error factors.

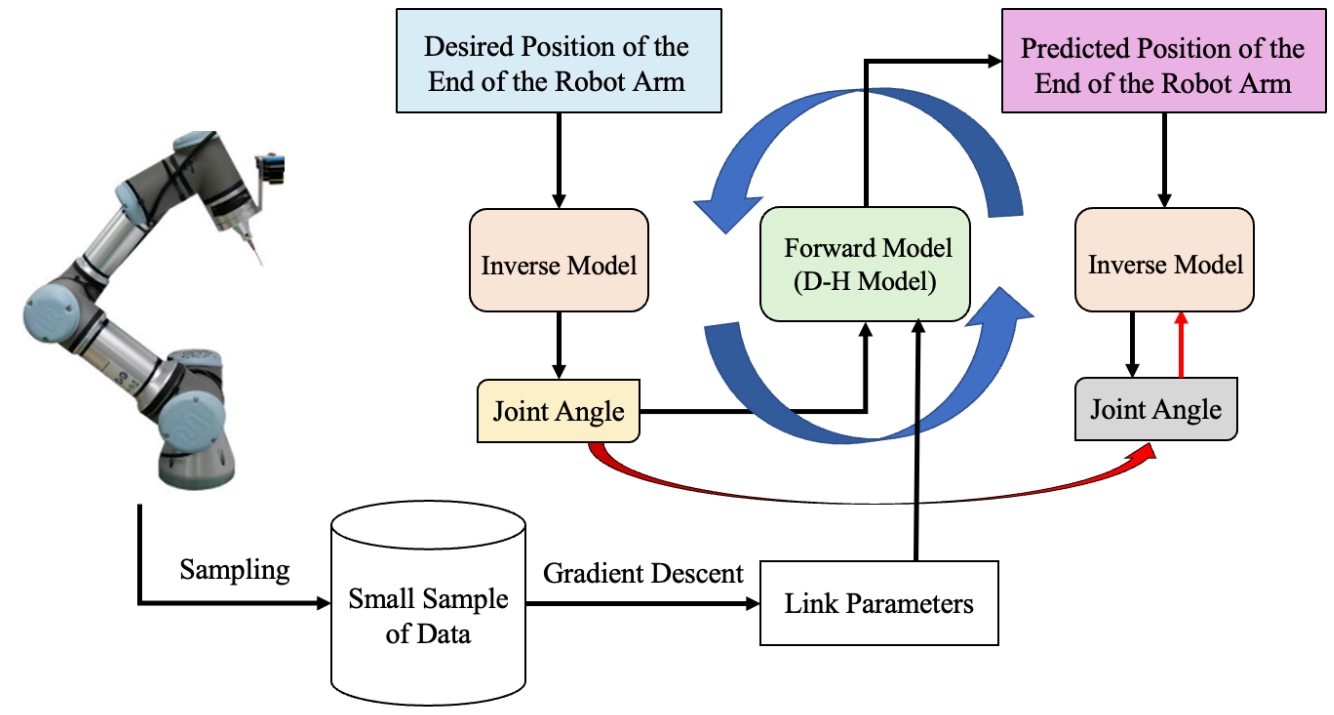

Fig.4. The framework for fast adaptation of robot arm models by sampling a small sample of data.

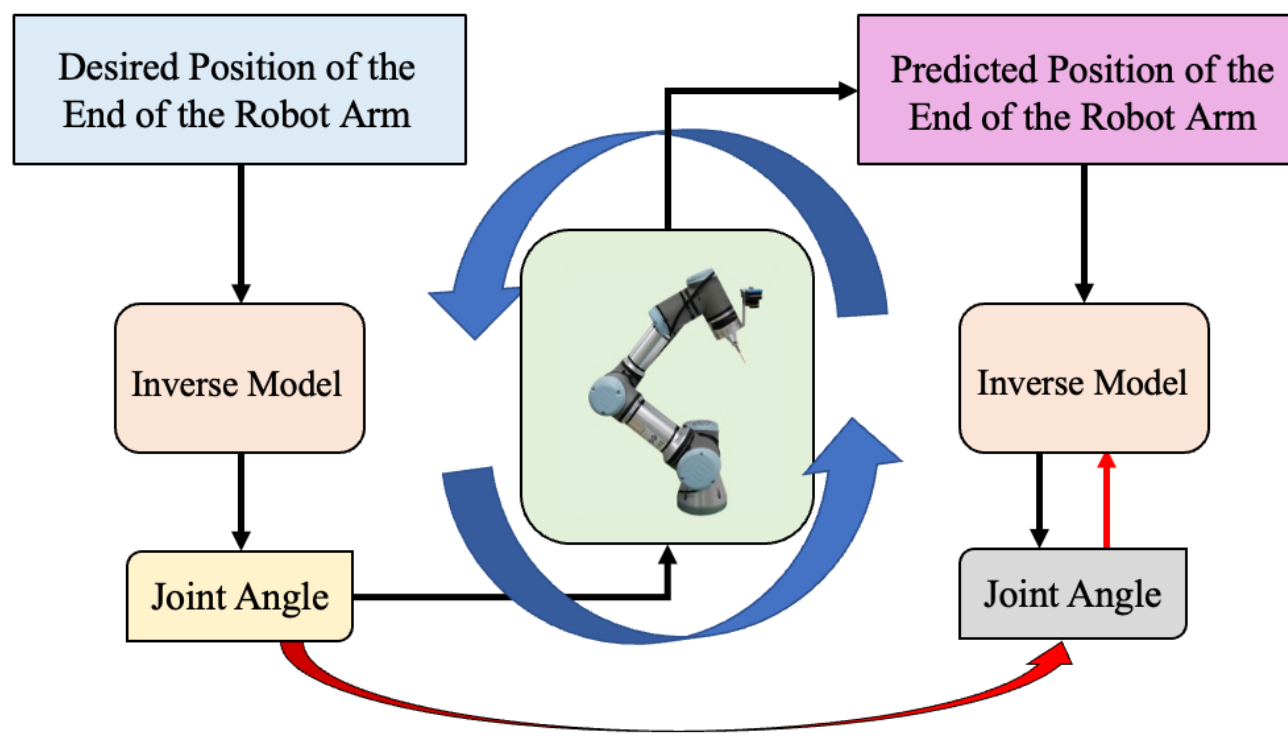

Fig.5. The framework for fast adaptation of robot arm models through the real robot arm.

## IV. Experiments

In this paper, we focus solely on the end-effector position, disregarding the end-effector orientation. As a result, the 6-DOF robot arm utilized in this work exhibits a redundancy of 3 degrees of freedom. Moreover, it is important to note that visual perception positioning errors in the real-world environment can negatively affect model performance. However, such effects are obviously not the focus of this paper. Therefore, a series of experiments were conducted mainly on a simulation platform to investigate the performance of the method without visual perception positioning errors. As for the dataset, joint angles were obtained through random sampling in the joint space, and the end position was calculated using the forward model to construct the dataset. A total of 100,000 data were collected, with 70,000 used for training and 30,000 for testing.

### *A. Evaluation on Inverse Model EMSSL*

A neural network is used to represent the inverse model, whose hyperparameters are shown in Table I. The input and output of the inverse model are normalized. The evaluation metric used for the inverse model learning is the predicted end position distance error, which measures the accuracy of the inverse model in predicting the end position. Direct Regression Learning, Distal Supervised Learning and Conditional Generative Adversarial Network are used as baseline methods.

TABLE I. Hyperparameters Settings

| **Hyperparameters** | **Value** |
|---|---|
| Activation | ReLU (Hidden Layers)<br>Sigmoid (Output Layers) |
| Optimizer | Adam |
| Learning rate | 0.0015 |
| Batch size (Inference) | 512 |
| Batch size (Training) | 128 |
| Number of parallel computation threads | 6 |
| Maximum number of iterations | 200 |
| Epoch | 10 |
| Number of network layers | 6 |
| Network Type | Fully connected neural network (FCNN) |
| Network Structure | 3 → 1024 → 512 →256 → 128 → 6 |

TABLE II. Performance Comparison of Different Methods

| **Method** | **Distance Error (cm)** | |
|---|---|---|
| | **Joint Angle Space[-90°,90°]** | **Joint Angle Space[-180°,180°]** |
| Direct Regression Learning (DRL) | 2.34 | 31.69 |
| Distal Supervised Learning (DSL) | 0.39 | 2.37 |
| Conditional Generative Adversarial Network (CGAN) | 0.48 | 2.57 |
| **Embodied Self-supervised Learning (EMSSL)** | **0.28** | **1.78** |

The experimental results are presented in Table II, which shows that our proposed method outperforms other baseline methods. As is shown in Table II, the performance of direct regression learning (DRL) is the poorest, with a significant difference compared to the results of other methods, which further underscores the inability of direct regression learning to effectively handle the non-convex problem in inverse model learning. Distal supervised learning (DSL) can achieve relatively good performance, but it requires prior training of the forward model, and its ultimate performance is limited by the cumulative training errors of both the forward and inverse models. The performance of the Conditional Generative Adversarial Network (CGAN) is superior to that of DRL, but inferior to the other two methods. In fact, the CGAN here learns the joint distribution between the joint angles and the end position, which results in a generally lower model accuracy. As the joint angle space expands, the non-convexity problem becomes more severe, and all methods perform worse, with DRL being completely infeasible.

The proposed method shows close convergence after 20 iterations, as illustrated in Fig. 6, which also demonstrates that EMSSL with sampling and training coordination can effectively facilitate the learning of the inverse model.

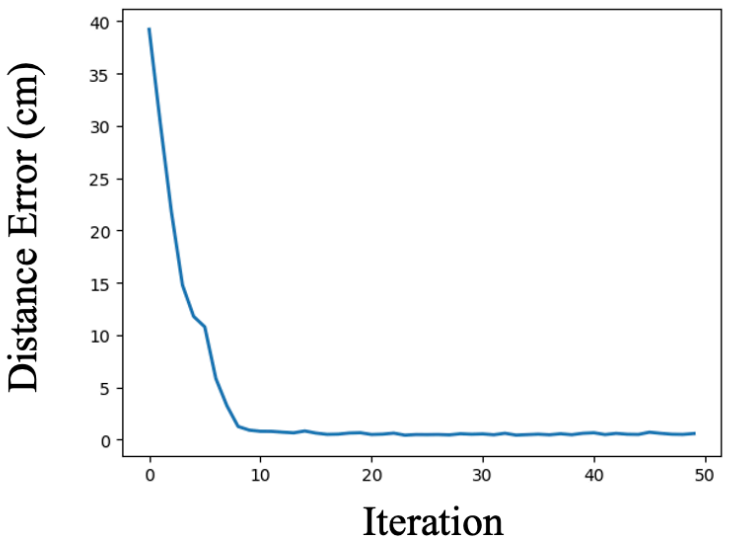

Fig.6. Learning performance curve of embodied self-supervised learning (EMSSL) with sampling and training coordination.

## B. Evaluation on Data Sampling Accelerating

In this paper, we conducted experiments on a computing platform equipped with a 6-core, 12-thread i7-8700 (3.2 GHz) CPU and a GeForce GTX 2080 (8 GB) GPU. We analyzed the time consumption of data sampling under different batch sizes and thread numbers. Fig. 7(a) shows that the time consumption of data sampling decreases as the batch size increases, reaching a minimum at a batch size of 512. Similarly, Fig. 7(b) illustrates that the sampling time decreases as the number of threads increases, with the minimum time consumption achieved at 6 threads, corresponding to the maximum number of CPU cores. These findings demonstrate that the optimal number of physical parallel threads depends on the number of CPU cores, and adding more threads beyond the number of cores may introduce additional overhead, such as thread switching.

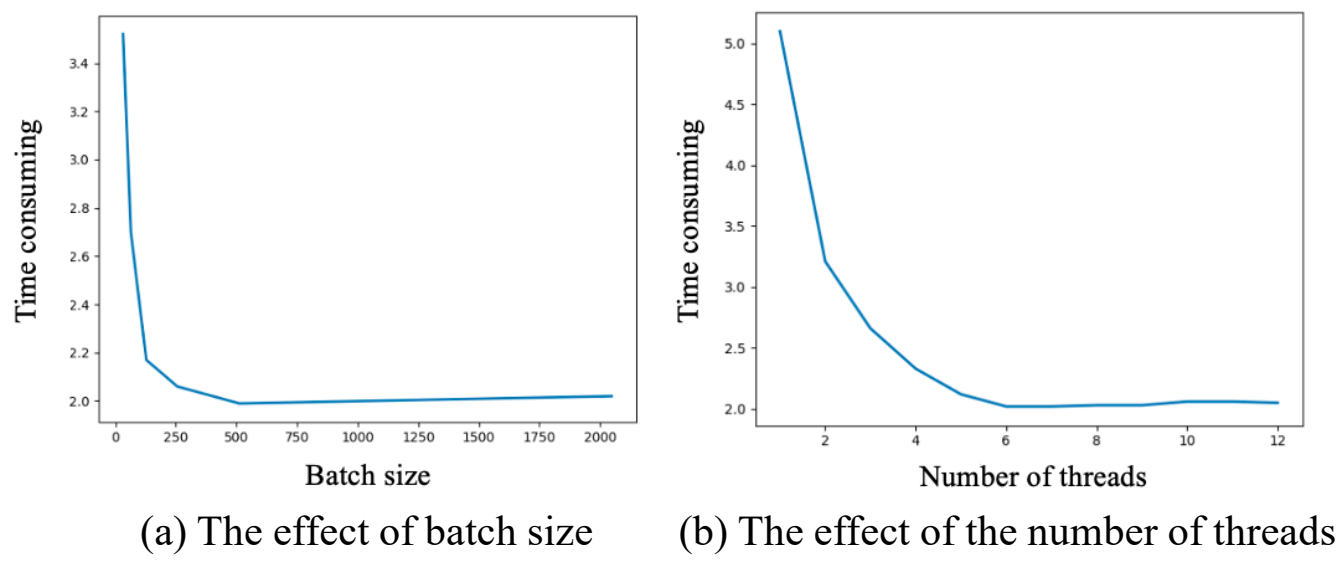

(a) The effect of batch size (b) The effect of the number of threads

Fig.7. Effect of batch size and the number of threads on the time consuming in the data sampling phase.

TABLE III. TIME CONSUMING OF DIFFERENT STRATEGIES

| Strategy | Time Consuming (s) |
|---|---|
| No Accelerating | 61.40 |
| Batch Inference Only | 5.10 |
| Parallel Computation Only | 57.84 |
| **Both Batch Inference and Parallel Computation** | **2.02** |

We also compared the time consumption of four different strategies: no acceleration, batch inference only, parallel computation only, and a combination of batch inference and parallel computation, with a batch size of 512 for batch inference and 6 threads for parallel computation. The results are shown in Table III, indicating that the strategy that combines batch inference and parallel computation has a significant speed-up effect on data sampling (up to 30 times faster).

## C. Evaluation on Fast Model Adaptation

To investigate the fast adaptation of the inverse model, we simulated changes in the length of the robot arm link. Specifically, we increased the length of each link by the same amount. As per the results in section IV-A, the distance error of the inverse model was 0.28 cm when the robot arm was not changed. The results in Table IV indicate that the initial error of the inverse model becomes larger when the robot arm length changes. However, after several iterations, the inverse model adapts to the changing length. Although the final error of the inverse model after adaptation increases as the length of the arm changes, the increment is relatively small. These findings demonstrate the effectiveness of the proposed approaches in achieving fast adaptation of the inverse model when the robot arm changes.

TABLE IV. RESULTS OF FAST ADAPTATION

| Length change (cm) | Distance Error (cm) | | Iteration |
|---|---|---|---|
| | Before Adaptation | After Adaptation | |
| 1.00 | 2.98 | 0.28 | **2** |
| 3.00 | 8.87 | 0.30 | **3** |
| 5.00 | 14.78 | 0.36 | **7** |

# V. CONCLUSION AND FUTURE WORK

In this paper, we propose a framework for autonomous learning of robot arm inverse kinematic model based on embodied self-supervised learning (EMSSL) with coordinated sampling and training. We emphasize the introduction of embodied cues for the robot, learning with the desired position of the end of the robot arm as the target, and solving the non-convex problem of the inverse model through the constraints of the forward model. Additionally, in contrast to current methods, the proposed method has an iterative and progressive relationship between data sampling and model training, rather than a sequential relationship. Our proposed method outperforms other baseline methods and has a faster convergence rate, with an average of 20 iterations. To address the slow data sampling phase, we investigate an optimized acceleration strategy based on inverse model batch inference and forward model parallel computation, which can significantly reduce the time required by up to 30 times. Furthermore, we develop two fast adaptation approaches under the framework of EMSS, drawing from the characteristics of human learning, and validate the proposed methods by simulating the change of robot arm length. Experimental results show the effectiveness of the proposed approaches. Specifically, the number of iterations required for the model adaptation is 2 if the length of each link changes 1 cm, and 3 if the length of each link changes 3 cm.

However, it is important to note that the experiments in this paper were mainly conducted on a simulation platform due to the effect of visual perception position errors in real-world environments. Further research is needed to reduce the effect of visual errors in order to realize the proposed method on a real robot arm.

## ACKNOWLEDGMENT

The work is supported in part by the National Natural Science Foundation of China (No. 62176004, No. U1713217), Intelligent Robotics and Autonomous Vehicle Lab (RAV), the Fundamental Research Funds for the Central Universities and High-performance Computing Platform of Peking University.

## REFERENCES

[1] Critchlow A J. Introduction to robotics[J]. 1985.

[2] Duy Nguyen-Tuong and Jan Peters. “Model learning for robot control: A survey”. Cognitive Pro- cessing, 2011, 12(4): 319–340.

[3] Yiannis Demiris and Anthony Dearden. “From motor babbling to hierarchical learning by imitation: A robot developmental pathway”. In: International Workshop on Epigenetic Robotics. 2005: 31–37.

[4] Daniele Caligiore, Tomassino Ferrauto, Domenico Parisi et al. “Using motor babbling and hebb rules for modeling the development of reaching with obstacles and grasping”. In: International Conference on Cognitive Systems (ICCS). 2008: E1–8.

[5] Matthias Rolf, Jochen J Steil and Michael Gienger. “Goal babbling permits direct learning of inverse kinematics”. IEEE Transactions on Autonomous Mental Development, 2010, 2(3): 216–229.

[6] Matthias Rolf, Jochen J Steil and Michael Gienger. “Online goal babbling for rapid bootstrapping of inverse models in high dimensions”. In: Joint IEEE International Conference on Development and Learning and Epigenetic Robotics (ICDL-EpiRob). 2011: 1–8.

[7] Erhard Wieser and Gordon Cheng. “Predictive action selector for generating meaningful robot behaviour from minimum amount of samples”. In: Joint IEEE International Conference on Devel- opment and Learning and Epigenetic Robotics (ICDL-EpiRob). 2014: 139–145.

[8] Erhard Wieser and Gordon Cheng. “Progressive learning of sensory-motor maps through spa- tiotemporal predictors”. In: Joint IEEE International Conference on Development and Learning and Epigenetic Robotics (ICDL-EpiRob). 2016: 43–48.

[9] Yifan Sun and Xihong Wu. “Embodied self-supervised learning by coordinated sampling and training”. arXiv preprint arXiv:2006.13350, 2020.

[10] Robert E Schapire and Yoav Freund. “Boosting: Foundations and algorithms”. Kybernetes, 2013.

[11] H Peyton Young. “Learning by trial and error”. Games and Economic Behavior, 2009, 65(2): 626–643.

[12] Rosen Diankov. Automated construction of robotic manipulation programs [phdthesis]. Carnegie Mellon University, 2010.

[13] Samuel R Buss. “Introduction to inverse kinematics with jacobian transpose, pseudoinverse and damped least squares methods”. IEEE Journal of Robotics and Automation, 2004, 17(1-19): 16.

[14] Patrick Beeson and Barrett Ames. “TRAC-IK: An open-source library for improved solving of generic inverse kinematics”. In: IEEE-RAS International Conference on Humanoid Robots (Humanoids). 2015: 928–935.

[15] Ahmed RJ Almusawi, L Canan Dülger and Sadettin Kapucu. “A new artificial neural network approach in solving inverse kinematics of robotic arm (denso vp6242)”. Computational Intelligence and Neuroscience, 2016.

[16] Hassan Ashraf Elkholy, Abdalla Saber Shahin, Abdelaziz Wasfy Shaarawy et al. “Solving inverse kinematics of a 7-DOF manipulator using convolutional neural network”. In: Joint European-US Workshop on Applications of Invariance in Computer Vision. 2020: 343–352.

[17] S Phaniteja, Parijat Dewangan, Pooja Guhan et al. “A deep reinforcement learning approach for dynamically stable inverse kinematics of humanoid robots”. In: IEEE International Conference on Robotics and Biomimetics (ROBIO). 2017: 1818–1823.

[18] Dmitriy Blinov, Anton Saveliev and Aleksandra Shabanova. “Deep q-learning algorithm for solving inverse kinematics of four-link manipulator”. In: International Conference on Electromechanics and Robotics "Zavalishin’s Readings". 2021: 279–291.

[19] Malik A, Lischuk Y, Henderson T, et al. A Deep Reinforcement-Learning Approach for Inverse Kinematics Solution of a High Degree of Freedom Robotic Manipulator[J]. Robotics, 2022, 11(2): 44.

[20] Joey K Parker, Ahmad R Khoogar and David E Goldberg. “Inverse kinematics of redundant robots using genetic algorithms”. In: IEEE International Conference on Robotics and Automation (ICRA). 1989: 271–272.

[21] Shaher Momani, Zaer S Abo-Hammour and Othman MK Alsmadi. “Solution of inverse kinematics problem using genetic algorithms”. Applied Mathematics & Information Sciences, 2016, 10(1): 225.

[22] Sebastian Starke, Norman Hendrich, Sven Magg et al. “An efficient hybridization of genetic algo- rithms and particle swarm optimization for inverse kinematics”. In: IEEE International Conference on Robotics and Biomimetics (ROBIO). 2016: 1782–1789.

[23] Michael I Jordan and David E Rumelhart. “Forward models: Supervised learning with a distal teacher”. Cognitive Science, 1992, 16(3): 307–354.

[24] Aaron D’Souza, Sethu Vijayakumar and Stefan Schaal. “Learning inverse kinematics”. In: IEEE/RSJ International Conference on Intelligent Robots and Systems (IROS). 2001: 298–303.

[25] Sethu Vijayakumar and Stefan Schaal. “Locally weighted projection regression: An o (n) algorithm for incremental real time learning in high dimensional space”. In: International Conference on Machine Learning (ICML). 2000: 288–293.

[26] Matthias Rolf and Jochen J Steil. “Efficient exploratory learning of inverse kinematics on a bionic elephant trunk”. IEEE Transactions on Neural Networks and Learning Systems, 2013, 25(6): 1147– 1160.

[27] Wolfram Schenck. Adaptive internal models for motor control and visual prediction. Logos Verlag Berlin GmbH, 2008.

[28] Tim von Oehsen, Alexander Fabisch, Shivesh Kumar et al. “Comparison of distal teacher learning with numerical and analytical methods to solve inverse kinematics for rigid-body mechanisms”. arXiv preprint arXiv:2003.00225, 2020.

[29] Christopher M Bishop. Mixture density networks [techreport]. 1994.

[30] Jakob Kruse, Lynton Ardizzone, Carsten Rother et al. “Benchmarking invertible architectures on inverse problems”. In: ICML Workshop on Invertible Neural Networks and Normalizing Flows. 2019.

[31] Lynton Ardizzone, Jakob Kruse, Carsten Rother et al. “Analyzing inverse problems with invertible neural networks”. In: International Conference on Learning Representations (ICLR). 2019.

[32] Hailin Ren and Pinhas Ben-Tzvi. “Learning inverse kinematics and dynamics of a robotic ma- nipulator using generative adversarial networks”. Robotics and Autonomous Systems, 2020, 124: 103386.

[33] Teguh Santoso Lembono, Emmanuel Pignat, Julius Jankowski et al. “Learning constrained distribu- tions of robot configurations with generative adversarial network”. IEEE Robotics and Automation Letters, 2021, 6(2): 4233–4240.

[34] Mehdi Mirza and Simon Osindero. “Conditional generative adversarial nets”. arXiv preprint arXiv:1411.1784, 2014.

[35] Clément Moulin-Frier and Pierre-Yves Oudeyer. “Exploration strategies in developmental robotics: A unified probabilistic framework”. In: Joint IEEE International Conference on Development and Learning and Epigenetic Robotics (ICDL-EpiRob). 2013: 1–6.

[36] Adrien Baranes and Pierre-Yves Oudeyer. “Active learning of inverse models with intrinsically motivated goal exploration in robots”. Robotics and Autonomous Systems, 2013, 61(1): 49–73.